# A Fusion Algorithm for Solving Bayesian Decision Problems

Prakash P. Shenoy
School of Business
University of Kansas
Lawrence, KS 66045-2003

## Abstract

This paper proposes a new method for solving Bayesian decision problems. The method consists of representing a Bayesian decision problem as a valuation-based system and applying a fusion algorithm for solving it. The fusion algorithm is a hybrid of local computational methods for computation of marginals of joint probability distributions and the local computational methods for discrete optimization problems.

## 1 INTRODUCTION

The main goal of this paper is to describe a new method for solving Bayesian decision problems. The method consists of representing a Bayesian decision problem as a valuation-based system and applying a fusion algorithm for solving it.

Valuation-based systems are described in Shenoy [1989, 1991c]. In valuation-based system representations of decision problems, we encode utility functions and probability distributions by functions called valuations. We solve valuation-based systems using two operations called combination and marginalization. Solving can be described simply as marginalizing all variables out of the joint valuation. The joint valuation is the result of combining all valuations. The framework of valuation-based systems is powerful enough to include also probability theory [Shenoy, 1991c], Dempster-Shafer theory of belief functions [Shenoy, 1991c], Spohn's theory of epistemic beliefs [Shenoy, 1991a,c], possibility theory [Dubois and Prade, 1990], discrete optimization [Shenoy, 1991b], propositional logic [Shenoy, 1990a], and constraint satisfaction problems [Shenoy and Shafer, 1988].

The fusion algorithm for solving valuation-based representations of decision problems is a hybrid of local computational methods for computation of marginals of joint probability distributions and local computational methods for discrete optimization. Local computational methods for computation of marginals of joint probability distributions have been proposed by, e.g., Pearl [1988], Lauritzen and Spiegelhalter [1988], Shafer and Shenoy [1988], and Jensen et al. [1990]. Local computational methods for discrete optimization are also called non-serial dynamic programming [Bertele and Brioschi, 1972]. Viewed abstractly using the framework of valuation-based systems, these two local computational methods are actually similar. Shenoy and Shafer [1990] and Shenoy [1991b] show that the same three axioms justify the use of local computation in both these cases.

Our method for representing and solving decision problems has many similarities to influence diagram methodology [Howard and Matheson, 1984; Olmsted, 1983; Shachter, 1986; Ezawa, 1986; Tatman, 1986]. But there are also many differences both in representation and solution. A comparison of these two methods is given in [Shenoy, 1990b].

We describe our new method using a diabetes diagnosis problem. Section 2 gives a statement of this problem. Section 3 describes a valuation-based representation of a decision problem. Section 4 describes the method for solving valuation-based systems. Section 5 describes a fusion algorithm for solving valuation-based systems using local computation. Finally, section 6 summarizes the paper.

## 2 A DIABETES DIAGNOSIS PROBLEM

A medical intern is trying to decide on a policy for treating patients suspected of suffering from diabetes. The intern first observes whether a patient exhibits two symptoms of diabetes—blue toe and glucose in urine. After she observes the presence or absence of these symptoms, she then either prescribes a treatment for diabetes or doesn't.

Table 1 shows the intern's utility function. Also, for the population of patients served by the intern, the prior probability of diabetes is 10%. Furthermore, for patients known to suffer from diabetes, 1.4% exhibit blue toe, and 90% exhibit glucose in urine. On the other hand, for patients known not to suffer from diabetes, 0.6% exhibit

Table 1. The Intern's Utility Function

| Intern's utilities ($\pi$) | | Act | |
|---|---|---|---|
| | | treat for diabetes (t) | not treat (~t) |
| State | has diabetes (d) | 10 | 0 |
| | no diabetes (~d) | 5 | 10 |

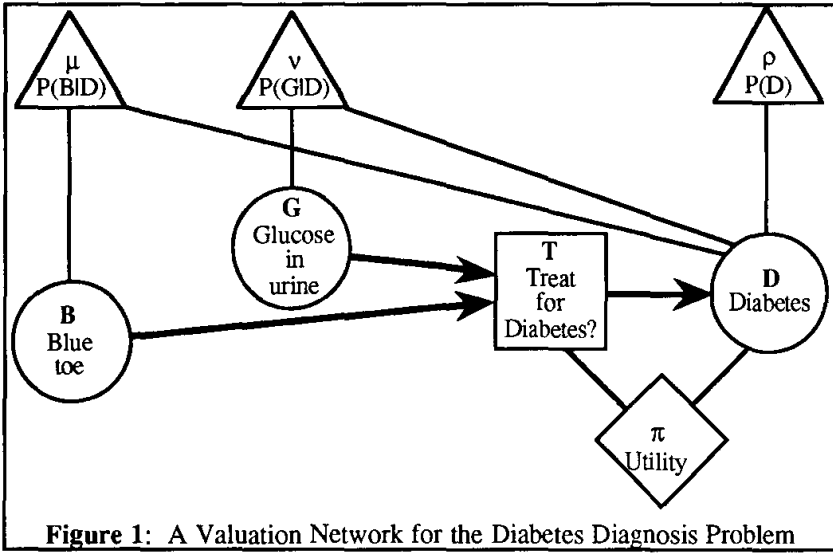

**Figure 1**: A Valuation Network for the Diabetes Diagnosis Problem

Let $\mathcal{X}_D$ denote the set of all decision variables, let $\mathcal{X}_R$ denote the set of all random variables, and let $\mathcal{X} = \mathcal{X}_D \cup \mathcal{X}_R$ denote the set of all variables. We will often deal with non-empty subsets of variables in $\mathcal{X}$. Given a non-empty subset h of $\mathcal{X}$, let $\mathcal{W}_h$ denote the Cartesian product of $\mathcal{W}_X$ for X in h, i.e., $\mathcal{W}_h = \times\{\mathcal{W}_X | X \in h\}$. We can think of the set $\mathcal{W}_h$ as the set of possible values of the joint variable h. Accordingly, we call $\mathcal{W}_h$ the *frame for h*. Also, we refer to elements of $\mathcal{W}_h$ as *configurations of h*. We use lower-case, bold-faced letters such as **x**, **y**, etc. to denote configurations. Also, if **x** is a configuration of g and **y** is a configuration of h and $g \cap h = \emptyset$, then (**x**,**y**) denotes a configuration of $g \cup h$.

blue toe, and 1% exhibit glucose in urine. We assume that blue toe and glucose in urine are conditionally independent given diabetes.

## 3 VALUATION-BASED SYSTEM REPRESENTATION

In this section, we describe a valuation-based system (VBS) representation of a decision problem. A VBS representation consists of decision variables, random variables, frames, a utility valuation, potentials, and precedence constraints. A graphical depiction of a VBS is called a *valuation network*. Figure 1 shows a valuation network for the diabetes diagnosis problem.

**Variables, Frames and Configurations.** A decision node is represented as a variable. The possible values of a decision variable represent the acts available at that point. We use the symbol $\mathcal{W}_D$ for the set of possible values of decision variable D. We assume that the decision-maker has to pick one and only one of the elements of $\mathcal{W}_D$ as their decision. We call $\mathcal{W}_D$ the *frame for D*. Decision variables are represented in valuation networks by rectangular nodes.

In the diabetes diagnosis problem, there is one decision node T. The frame for T has two elements: Treat the patient for diabetes (t), and not treat (~t).

If R is a random variable, we use the symbol $\mathcal{W}_R$ to denote its possible values. We assume that one and only one of the elements of $\mathcal{W}_R$ can be the true value of R. We call $\mathcal{W}_R$ the *frame for R*. Random variables are represented in valuation networks by circular nodes.

In the diabetes diagnosis problem, there are three random variables: Blue toe (B), Glucose in urine (G), and Diabetes (D). Each variable has a frame consisting of two elements.

It is convenient to extend this terminology to the case where the set of variables h is empty. We adopt the convention that the frame for the empty set $\emptyset$ consists of a single configuration, and we use the symbol ♦ to name that configuration; $\mathcal{W}_\emptyset = \{♦\}$. To be consistent with our notation above, we adopt the convention that if **x** is a configuration for g, then (**x**,♦) = **x**.

**Valuations.** Suppose $h \subseteq \mathcal{X}$. A *utility valuation* $\pi$ for h is a function from $\mathcal{W}_h$ to $\mathbb{R}$, where $\mathbb{R}$ denotes the set of real numbers. The values of utility valuations are utilities. If $h = d \cup r$ where $d \subseteq \mathcal{X}_D$ and $r \subseteq \mathcal{X}_R$, $\mathbf{x} \in \mathcal{W}_d$, and $\mathbf{y} \in \mathcal{W}_r$, then $\pi(\mathbf{x},\mathbf{y})$ denotes the utility to the decision maker if the decision maker chooses configuration **x** and the true configuration of r is **y**. If $\pi$ is a utility valuation for h, and $X \in h$, then we say that $\pi$ *bears on X*.

In a valuation network, a utility valuation is represented by a diamond-shaped node. To permit the identification of all valuations that bear on a variable, we draw undirected edges between the utility valuation node and all the variable nodes it bears on. In the diabetes diagnosis problem, there is one utility valuation $\pi$ as shown in Figure 1. Table 1 shows the values of this utility valuation.

Suppose $h \subseteq \mathcal{X}$. A *potential $\rho$ for h* is a function from $\mathcal{W}_h$ to the unit interval [0, 1]. The values of potentials are probabilities.

In a valuation network, a potential is represented by a triangular node. Again, to identify the variables related by a potential, we draw undirected edges between the potential node and all the variable nodes it bears on.

In the diabetes diagnosis problem, there are three poten-tials $\mu$, $\nu$, and $\rho$ as shown in Figure 1. Table 2 shows the details of these potentials. Note that $\mu$ is a potential for $\{B, D\}$, $\nu$ is a potential for $\{G, D\}$, and $\rho$ is a potential for $\{D\}$.

**Table 2**: Potentials $\rho$, $\mu$, and $\nu$

| D | $\rho$ |
|---|---|
| d | .1 |
| ~d | .9 |

| $\mu$ | | B | |
|---|---|---|---|
| | | b | ~b |
| D | d | .014 | .986 |
| | ~d | .006 | .994 |

| $\nu$ | | G | |
|---|---|---|---|
| | | b | ~b |
| D | d | .90 | .10 |
| | ~d | .01 | .99 |

**Precedence Constraints.**
Besides acts, states, probabilities and utilities, an important ingredient of problems in decision analysis is information constraints. Some decisions have to be made before the observation of some uncertain states, and some decisions can be postponed until after some states are observed. In the diabetes diagnosis problem, for example, the medical intern doesn't know whether the patient has diabetes or not. And the decision whether to treat the patient for diabetes or not may be postponed until after the observation of blue toe and glucose in urine.

If a decision-maker expects to be informed of the true value of random variable R before they make a decision D, then we represent this situation by the binary relation R→D (read as *R precedes D*). On the other hand, if a random variable R is only revealed after a decision D is made or perhaps never revealed, then we represent this situation by the binary relation D→R.

In the diabetes diagnosis problem, we have the precedence constraints B→T, G→T, T→D. The decision whether to treat the patient for diabetes or not (T) is only made after observing blue toe (B) and glucose in urine (G). And, diabetes (D) is not known at the time the decision whether to treat the patient for diabetes (T) has to be made.

Suppose > is a binary relation on $\mathcal{X}$ such that it is the transitive closure of →, i.e., X > Y if either X → Y, or there exists a Z∈ $\mathcal{X}$ such that X > Z and Z > Y. First, we assume that > is a partial order on $\mathcal{X}$ (otherwise the decision problem is ill-defined and not solvable). Second, we require that this partial order > is such that for any D∈ $\mathcal{X}_D$ and any R∈ $\mathcal{X}_R$, either D>R or R>D. We refer to this second condition as the *perfect recall condition*. The reason for the perfect recall condition is as follows. Given the meaning of the precedence relation →, for any decision variable D and any random variable R, either R is known when decision D has to be made, or not. This translates to either R>D or D>R. (This condition is called "no-forgetting assumption" in influence diagram literature.)

Next, we will define two operations called combination and marginalization. We use these operations to solve the valuation-based system representation. First we start with some notation.

**Projection of Configurations.** *Projection* of configurations simply means dropping extra coordinates; if

**Table 3:** Computation of $\mu \otimes \nu \otimes \rho$ and the Joint Valuation $\pi \otimes \mu \otimes \nu \otimes \rho$

| $\mathcal{W}_{\{B,G,T,D\}}$ | $\pi$ | $\mu$ | $\nu$ | $\rho$ | $\mu \otimes \nu \otimes \rho$ | $\pi \otimes \mu \otimes \nu \otimes \rho$ |
|---|---|---|---|---|---|---|
| b  g  t  d   | 10 | .014 | .90 | .10 | .00126 | 0.0126 |
| b  g  t  ~d  | 5  | .006 | .01 | .90 | .000054 | 0.00027 |
| b  g  ~t  d  | 0  | .014 | .90 | .10 | .00126 | 0 |
| b  g  ~t  ~d | 10 | .006 | .01 | .90 | .000054 | 0.00054 |
| b  ~g  t  d  | 10 | .014 | .10 | .10 | .00014 | 0.0014 |
| b  ~g  t  ~d | 5  | .006 | .99 | .90 | .005346 | 0.02673 |
| b  ~g  ~t  d | 0  | .014 | .10 | .10 | .00014 | 0 |
| b  ~g  ~t  ~d| 10 | .006 | .99 | .90 | .005346 | 0.05346 |
| ~b  g  t  d  | 10 | .986 | .90 | .10 | .08874 | 0.8874 |
| ~b  g  t  ~d | 5  | .994 | .01 | .90 | .008946 | 0.04473 |
| ~b  g  ~t  d | 0  | .986 | .90 | .10 | .08874 | 0 |
| ~b  g  ~t  ~d| 10 | .994 | .01 | .90 | .008946 | 0.08646 |
| ~b  ~g  t  d | 10 | .986 | .10 | .10 | .00986 | 0.0986 |
| ~b  ~g  t  ~d| 5  | .994 | .99 | .90 | .885654 | 4.42827 |
| ~b  ~g  ~t  d| 0  | .986 | .10 | .10 | .00986 | 0 |
| ~b  ~g  ~t  ~d| 10 | .994 | .99 | .90 | .885654 | 8.86554 |

(w,x,y,z) is a configuration of $\{W,X,Y,Z\}$, for example, then the projection of (w,x,y,z) to $\{W,X\}$ is simply (w,x), which is a configuration of $\{W,X\}$.

If g and h are sets of variables, h⊆g, and x is a configuration of g, then we let $x^{\downarrow h}$ denote the projection of x to h. The projection $x^{\downarrow h}$ is always a configuration of h. If h=g and x is a configuration of g, then $x^{\downarrow h}$ = x. If h=∅, then $x^{\downarrow h}$ = ♦.

**Combination.** The definition of combination depends on the type of valuations being combined.

Suppose h and g are subsets of $\mathcal{X}$, suppose $\rho_i$ is a potential for h, and suppose $\rho_j$ is a potential for g. Then the *combination of $\rho_i$ and $\rho_j$*, denoted by $\rho_i \otimes \rho_j$, is a potential for h∪g obtained by pointwise multiplication of $\rho_i$ and $\rho_j$, i.e., $(\rho_i \otimes \rho_j)(x) = \rho_i(x^{\downarrow h})\rho_j(x^{\downarrow g})$ for all $x \in \mathcal{W}_{h \cup g}$. See Table 3 for an example.

Suppose h and g are subsets of $\mathcal{X}$, suppose $\pi_i$ is a utility valuation for h, and suppose $\rho_j$ is a potential for g. Then the *combination of $\pi_i$ and $\rho_j$*, denoted by $\pi_i \otimes \rho_j$, is a utility valuation for h∪g obtained by pointwise multiplication of $\pi_i$ and $\rho_j$, i.e., $(\pi_i \otimes \rho_j)(x) = \pi_i(x^{\downarrow h})\rho_j(x^{\downarrow g})$ for all $x \in \mathcal{W}_{h \cup g}$. See Table 3 for an example.

Note that combination is commutative and associative. Thus, if $\{\alpha_1, ..., \alpha_k\}$ is a set of valuations, we write $\otimes\{\alpha_1, ..., \alpha_k\}$ to mean the combination of valuations in $\{\alpha_1, ..., \alpha_k\}$ in some sequence.

**Marginalization.**
Suppose h is a subset of variables and suppose α is a valuation for h. Marginalization is an operation where we reduce valuation α to a valuation α↓(h−{X}) for h−{X}. α↓(h−{X}) is called the *marginal of α for h−{X}*. Unlike combination, the definition of marginalization does not depend on the nature of α. But the definition of marginalization does depend on whether X is a decision or a random variable.

If R is a random variable, α↓(h−{R}) is obtained by summing α over the frame for R, i.e., α↓(h−{R})(c) = Σ{α(c,r)| r∈ 𝒲$_R$} for all c∈ 𝒲$_{h−\{R\}}$. Here, α could be either a utility valuation or a potential. See Table 4 for an example.

**Table 4:** The Computation of $\tau^{\downarrow\{B,G,T\}}$, $\tau^{\downarrow\{B,G\}}$, $\Psi_T$, $\tau^{\downarrow\{B\}}$, and $\tau^{\downarrow\varnothing}(\blacklozenge)$

| 𝒲$_{\{B,G,T,D\}}$ | | | | τ | $\tau^{\downarrow\{B,G,T\}}$ | $\tau^{\downarrow\{B,G\}}$ | $\Psi_T$ | $\tau^{\downarrow\{B\}}$ | $\tau^{\downarrow\varnothing}(\blacklozenge)$ |
|---|---|---|---|---|---|---|---|---|---|
| b | g | t | d | 0.0126 | 0.01287 | 0.01287 | t | 0.06633 | 9.864 |
| b | g | t | ~d | 0.00027 | | | | | |
| b | g | ~t | d | 0 | 0.00054 | | | | |
| b | g | ~t | ~d | 0.00054 | | | | | |
| b | ~g | t | d | 0.0014 | 0.02813 | 0.05346 | ~t | | |
| b | ~g | t | ~d | 0.02673 | | | | | |
| b | ~g | ~t | d | 0 | 0.05346 | | | | |
| b | ~g | ~t | ~d | 0.05346 | | | | | |
| ~b | g | t | d | 0.8874 | 0.93213 | 0.93213 | t | 9.79767 | |
| ~b | g | t | ~d | 0.04473 | | | | | |
| ~b | g | ~t | d | 0 | 0.08646 | | | | |
| ~b | g | ~t | ~d | 0.08646 | | | | | |
| ~b | ~g | t | d | 0.0986 | 4.52687 | 8.86554 | ~t | | |
| ~b | ~g | t | ~d | 4.42827 | | | | | |
| ~b | ~g | ~t | d | 0 | 8.86554 | | | | |
| ~b | ~g | ~t | ~d | 8.86554 | | | | | |

(τ denotes the joint valuation π⊗μ⊗ν⊗ρ)

If D is a decision variable, α↓(h−{D}) is obtained by maximizing α over the frame for D, i.e., α↓(h−{D})(c) = MAX{α(c,d)| d∈ 𝒲$_D$} for all c∈ 𝒲$_{h−\{D\}}$. Here, α must be a utility valuation. See Table 4 for an example.

We now state three lemmas regarding the marginalization operation. Lemma 3.1 states that in marginalizing two decision variables out of a valuation, the order in which the variables are eliminated does not affect the result. Lemma 3.2 states a similar result for marginalizing two random variables out of a valuation. Lemma 3.3 states that in marginalizing a decision variable and a random variable out of a valuation, the order in which the two variables are eliminated may make a difference.

**Lemma 3.1.** Suppose h is a subset of 𝒳 containing decision variables $D_1$ and $D_2$, and suppose α is a utility valuation for h. Then
(α↓(h−{$D_1$}))↓(h−{$D_1,D_2$})(c) =
(α↓(h−{$D_2$}))↓(h−{$D_1,D_2$})(c) for all
c∈ 𝒲$_{h−\{D_1,D_2\}}$.

**Lemma 3.2.** Suppose h is a subset of 𝒳 containing random variables $R_1$ and $R_2$, and suppose α is a valuation for h. Then (α↓(h−{$R_1$}))↓(h−{$R_1,R_2$})(c) = (α↓(h−{$R_2$}))↓(h−{$R_1,R_2$})(c) for all c∈ 𝒲$_{h−\{R_1,R_2\}}$.

**Lemma 3.3.** Suppose h is a subset of 𝒳 containing decision variable D and random variable R, and suppose α is a utility valuation for h. Then
(α↓(h−{D}))↓(h−{R,D})(c) ≥
(α↓(h−{R}))↓(h−{R,D})(c) for all c∈ 𝒲$_{h−\{R,D\}}$.

It is clear from Lemma 3.3, that in marginalizing more than one variable out of a valuation, the order of elimination of the variables may make a difference. As we will see shortly, we need to marginalize all variables out of the joint valuation. What sequence should we use? This is where the precedence constraints come into play. We define marginalization such that variable Y is marginalized before X whenever X>Y. Here is a formal definition.

Suppose h and g are non-empty subsets of 𝒳 such that g is a proper subset of h, suppose α is a valuation for h, and suppose > is a partial order on 𝒳 satisfying the perfect recall condition. The *marginal of α for g with respect to the partial order* >, denoted by α↓g, is a valuation for g defined as follows: α↓g =
(((α↓(h−{$X_1$}))↓(h−{$X_1,X_2$}))...)↓(h−{$X_1,X_2,...,X_k$}) (3.1)
where h−g = {$X_1, ..., X_k$} and $X_1X_2...X_k$ is a sequence of variables in h−g such that with respect to the partial order >, $X_1$ is a minimal element of h−g, $X_2$ is a minimal element of h−g−{$X_1$}, etc.

The marginalization sequence $X_1X_2...X_k$ may not be

unique since > is only a partial order. But, since > satisfies the perfect recall condition, it is clear from Lemmas 3.1 and 3.2, that the definition of $\alpha^{\downarrow g}$ in (3.1) is well defined.

**Strategy.** The main objective in solving a decision problem is to compute an optimal strategy. What constitutes a strategy? Intuitively, a strategy is a choice of an act for each decision variable D as a function of configurations of random variables R such that R>D. Let Pr(D) = $\{R \in \mathcal{X}_R \mid R > D\}$. We refer to Pr(D) as the *predecessors of* D. Thus *a strategy* $\sigma$ is a collection of functions $\{\xi_D\}_{D \in \mathcal{X}_D}$ where $\xi_D: \mathcal{W}_{Pr(D)} \to \mathcal{W}_D$.

**Solution for a Variable.** Computing an optimal strategy is a matter of bookkeeping. Each time we marginalize a decision variable out of a utility valuation using maximization, we store a table of optimal values of the decision variable where the maximums are achieved. We can think of this table as a function. We call this function "a solution" for the decision variable. Suppose h is a subset of variables such that decision variable D∈ h, and suppose $\pi$ is a utility valuation for h. A function $\Psi_D: \mathcal{W}_{h-\{D\}} \to \mathcal{W}_D$ is called a *solution for D* (with respect to $\pi$) if $\pi^{\downarrow(h-\{D\})}(c) = \pi(c, \Psi(c))$ for all $c \in \mathcal{W}_{h-\{D\}}$. See Table 4 for an example.

## 4 SOLVING A VBS

Suppose $\Delta = \{\mathcal{X}_D, \mathcal{X}_R, \{\mathcal{W}_X\}_{X \in \mathcal{X}}, \{\pi_1\}, \{\rho_1, ..., \rho_n\}, \to\}$ is a VBS representation of a decision problem consisting of one utility valuation and n potentials. What do the potentials represent? And how do we solve $\Delta$? We will answer these two related questions in terms of a canonical decision problem.

**Canonical Decision Problem.** A *canonical decision problem* $\Delta_C$ consists of a single decision variable D with a finite frame $\mathcal{W}_D$, a single random variable R with a finite frame $\mathcal{W}_R$, a single utility valuation $\pi$ for $\{D,R\}$, a single potential $\rho$ for $\{R, D\}$ such that
$$\Sigma\{\rho(d,r) \mid r \in \mathcal{W}_R\} = 1 \text{ for all } d \in \mathcal{W}_D, \quad (4.1)$$
and a precedence relation $\to$ defined by D$\to$R.

The meaning of the canonical decision problem is as follows. The elements of $\mathcal{W}_D$ are acts, and the elements of $\mathcal{W}_R$ are states of nature. The potential $\rho$ is a family of probability distributions for R, one for each act $d \in \mathcal{W}_D$, i.e., $\Sigma\{\rho(d,r) \mid r \in \mathcal{W}_R\} = 1$ for all $d \in \mathcal{W}_D$.

The utility valuation $\pi$ is a utility function—if the decision maker chooses act d, and the state of nature r prevails, then the utility to the decision maker is $\pi(d,r)$. The precedence relation $\to$ states that the true state of nature is revealed to the decision maker only after the decision maker has chosen an act.

Solving a canonical decision problem using the criterion of maximizing expected utility is easy. The expected utility associated with act d is $\Sigma\{(\pi \otimes \rho)(d,r) \mid r \in \mathcal{W}_R\} = (\pi \otimes \rho)^{\downarrow\{D\}}(d)$. The maximum expected utility (associated with an optimal act, say $d^*$) is
$MAX\{(\pi \otimes \rho)^{\downarrow\{D\}}(d) \mid d \in \mathcal{W}_D\} = ((\pi \otimes \rho)^{\downarrow\{D\}})^{\downarrow\varnothing}(\blacklozenge) = (\pi \otimes \rho)^{\downarrow\varnothing}(\blacklozenge)$. Finally, act $d^*$ is optimal if and only if $(\pi \otimes \rho)^{\downarrow\{D\}}(d^*) = (\pi \otimes \rho)^{\downarrow\varnothing}(\blacklozenge)$.

Consider the decision problem $\Delta = \{\mathcal{X}_D, \mathcal{X}_R, \{\mathcal{W}_X\}_{X \in \mathcal{X}}, \{\pi_1\}, \{\rho_1, ..., \rho_n\}, \to\}$. We will explain the meaning of $\Delta$ by reducing it to an equivalent canonical decision problem $\Delta_C = \{\{D\}, \{R\}, \{\mathcal{W}_D, \mathcal{W}_R\}, \{\pi\}, \{\rho\}, \to\}$. To define $\Delta_C$, we need to define $\mathcal{W}_D, \mathcal{W}_R, \pi$, and $\rho$. Define $\mathcal{W}_D$ such that for each distinct strategy $\sigma$ of $\Delta$, there is a corresponding act $d_\sigma$ in $\mathcal{W}_D$. Define $\mathcal{W}_R$ such that for each distinct configuration y of $\mathcal{X}_R$ in $\Delta$, there is a corresponding configuration $r_y$ in $\mathcal{W}_R$.

Before we define utility valuation $\pi$ for $\{D,R\}$, we need some notation. Suppose $\sigma = \{\xi_D\}_{D \in \mathcal{X}_D}$ is a strategy, and suppose y is a configuration of $\mathcal{X}_R$. Then together $\sigma$ and y determine a unique configuration of $\mathcal{X}_D$. Let $a_{\sigma,y}$ denote this unique configuration of $\mathcal{X}_D$. By definition, $a_{\sigma,y}^{\downarrow\{D\}} = \xi_D(y^{\downarrow Pr(D)})$ for all $D \in \mathcal{X}_D$. Consider the utility valuation $\pi_1$ in $\Delta$. Assume that the domain of this valuation includes all of $\mathcal{X}_D$. Typically the domain of this valuation will include also some (or all) random variables. Let p denote the subset of random variables included in the domain of the joint utility valuation, i.e., $p \subseteq \mathcal{X}_R$ such that $\pi_1$ is a utility valuation for $\mathcal{X}_D \cup p$. Define utility valuation $\pi$ for $\{D,R\}$ such that $\pi(d_\sigma, r_y) = \pi_1(a_{\sigma,y}, y^{\downarrow p})$, for all strategy $\sigma$ of $\Delta$, and all configuration $y \in \mathcal{W}_{\mathcal{X}_R}$. Remember that $a_{\sigma,y}$ is the unique configuration of $\mathcal{X}_D$ determined by $\sigma$ and y.

Consider the joint potential $\rho_1 \otimes ... \otimes \rho_n$. Assume that this potential includes all random variables in its domain. Let q denote the subset of decision variables included in the domain of the joint potential, i.e., $q \subseteq \mathcal{X}_D$ such that $\rho_1 \otimes ... \otimes \rho_n$ is a potential for $q \cup \mathcal{X}_R$. Note that q could be empty. Define potential $\rho$ for $\{D,R\}$ such that $\rho(d_\sigma, r_y) = (\rho_1 \otimes ... \otimes \rho_n)(a_{\sigma,y}^{\downarrow q}, y)$, for all strategy $\sigma$, and all configuration $y \in \mathcal{W}_{\mathcal{X}_R}$. $\Delta_C$, as defined above, is a canonical decision problem only if $\rho$ satisfies condition (4.1). This motivates the following definition. $\Delta$ is a *well-defined VBS representation of a decision problem* if and only if $\Sigma\{(\rho_1 \otimes ... \otimes \rho_n)(x,y) \mid y \in \mathcal{W}_{\mathcal{X}_R}\} = 1$ for every $x \in \mathcal{W}_q$.

In summary, the potentials $\{\rho_1, ..., \rho_n\}$ represent the factors of a family of probability distributions. It is easy to verify that the VBS representation of the diabetes diagnosis problem is well-defined since $\rho \otimes \mu \otimes \nu$ is a joint probability distribution for $\{D, B, G\}$.

**The Decision Problem.** Suppose $\Delta = \{\mathcal{X}_D, \mathcal{X}_R, \{\mathcal{W}_X\}_{X \in \mathcal{X}}, \{\pi_1\}, \{\rho_1, ..., \rho_n\}, \to\}$ is a well-defined deci-sion problem. Let $\Delta_C = \{\{D\}, \{R\}, \{\mathcal{W}_D, \mathcal{W}_R\}, \{\pi\}, \{\rho\}, \rightarrow\}$, represent an equivalent canonical decision problem. In the canonical decision problem $\Delta_C$, the two computations that are of interest are (1) the computation of the maximum expected value $(\pi \otimes \rho)^{\downarrow \varnothing}(\blacklozenge)$, and (2) the computation of an optimal act $\mathbf{d}_{\sigma^*}$ such that $(\pi \otimes \rho)^{\downarrow \{D\}}(\mathbf{d}_{\sigma^*}) = (\pi \otimes \rho)^{\downarrow \varnothing}(\blacklozenge)$. Since we know the mapping between $\Delta$ and $\Delta_C$, we can now formally define the questions posed in a decision problem $\Delta$. There are two computations of interest.

First, we would like to compute the maximum expected utility. The maximum expected utility is given by $(\otimes \{\pi_1, \rho_1, ..., \rho_n\})^{\downarrow \varnothing}(\blacklozenge)$. Second, we would like to compute an optimal strategy $\sigma^*$ that gives us the maximum expected value $(\otimes \{\pi_1, \rho_1, ..., \rho_n\})^{\downarrow \varnothing}(\blacklozenge)$. A strategy $\sigma^*$ of $\Delta$ is *optimal* if $(\pi \otimes \rho)^{\downarrow \{D\}}(\mathbf{d}_{\sigma^*}) = (\otimes \{\pi_1, \rho_1, ..., \rho_n\})^{\downarrow \varnothing}(\blacklozenge)$, where $\pi$, $\rho$, and D refer to the equivalent canonical decision problem $\Delta_C$.

In the diabetes diagnosis problem, we have four valuations $\pi$, $\mu$, $\nu$, and $\rho$. Also, from the precedence constraints, we have B>T, G>T, T>D. Thus we need to compute either $((((\pi \otimes \mu \otimes \nu \otimes \rho)^{\downarrow \{B,G,T\}})^{\downarrow \{B,G\}})^{\downarrow \{B\}})^{\downarrow \varnothing}$ or $((((\pi \otimes \mu \otimes \nu \otimes \rho)^{\downarrow \{B,G,T\}})^{\downarrow \{B,G\}})^{\downarrow \{G\}})^{\downarrow \varnothing}$. In either case, we get the same answer.

Tables 3 and 4 display the former computations. As seen in Table 4, the maximum expected utility is 9.864. Also, from $\Psi_T$, the solution for T (shown in Table 4), the optimal act is to treat the patient for diabetes if and only if the patient exhibits glucose in urine.

Note that no divisions were done in the solution process, only additions and multiplications. But, both decision tree and influence diagram methodologies involve unnecessary divisions, and unnecessary multiplications to compensate for the unnecessary divisions. It is this feature of valuation-based systems that makes it more efficient than decision trees and influence diagrams.

In solving the diabetes diagnosis problem using our method, we do only 11 additions, 28 multiplications and 4 comparisons, for a total of 43 operations. On the other hand, both decision tree and influence diagram methodologies require 17 additions, 38 multiplications, 12 divisions, and 4 comparisons for a total of 71 operations. Thus, for this problem, our method results in a savings of 40 percent over the decision tree and influence diagram methodologies.

## 5 A FUSION ALGORITHM

In this section, we describe a method for solving a VBS using local computation. The solution for the diabetes diagnosis problem shown in Tables 3 and 4 involves combination on the space $\mathcal{W}_{\mathcal{X}}$. While this is possible for small problems, it is computationally not feasible for problems with many variables. Given the structure of the diabetes diagnosis problem, it is not possible to avoid the combination operation on the space of all four variables, B, G, T, and D. But, in some problems, it may be possible to avoid such global computations.

The basic idea of the method is to successively delete all variables from the VBS. The sequence in which variables are deleted must respect the precedence constraints in the sense that if X>Y, then Y must be deleted before X. Since > is only a partial order, a problem may allow several deletion sequences. Any allowable deletion sequence may be used. All allowable deletion sequences lead to the same answers. But, different deletion sequences may involve different computational costs. We will comment on good deletion sequences at the end of this section.

When we delete a variable, we have to do a "fusion" operation on the valuations. Consider a set of k valuations $\alpha_1, ..., \alpha_k$. Suppose $\alpha_i$ is a valuation for $h_i$. Let $\text{Fus}_X\{\alpha_1, ..., \alpha_k\}$ denote the collection of valuations after fusing the valuations in the set $\{\alpha_1, ..., \alpha_k\}$ with respect to variable X. Then

$$\text{Fus}_X\{\alpha_1, ..., \alpha_k\} = \{\alpha^{\downarrow(h-\{X\})}\} \cup \{\alpha_i \mid X \notin h_i\}, \quad (5.1)$$

where $\alpha = \otimes\{\alpha_i \mid X \in h_i\}$, and $h = \cup\{h_i \mid X \in h_i\}$. After fusion, the set of valuations is changed as follows. All valuations that bear on X are combined, and the resulting valuation is marginalized such that X is eliminated from its domain. The valuations that do not bear on X remain unchanged.

We are ready to state the main theorem.

**Theorem 1.** Suppose $\Delta = \{\mathcal{X}_D, \mathcal{X}_R, \{\mathcal{W}_X\}_{X \in \mathcal{X}}, \{\pi_1\}, \{\rho_1, ..., \rho_n\}, \rightarrow\}$ is a well-defined decision problem. Suppose $X_1 X_2 ... X_k$ is a sequence of variables in $\mathcal{X} = \mathcal{X}_D \cup \mathcal{X}_R$ such that with respect to the partial order >, $X_1$ is a minimal element of $\mathcal{X}$, $X_2$ is a minimal element of $\mathcal{X} - \{X_1\}$, etc. Then $\{(\otimes\{\pi_1, \rho_1, ..., \rho_n\})^{\downarrow \varnothing}\} = \text{Fus}_{X_k}\{...\text{Fus}_{X_2}\{\text{Fus}_{X_1}\{\pi_1, \rho_1, ..., \rho_n\}\}\}$.

See [Shenoy, 1990c] for a proof. To illustrate Theorem 1, consider a VBS for a medical diagnosis problem as shown in Figure 2. In this VBS, there are three random variables, D, P, and S, and one decision variable, T. D represents a disease, P represents a pathological state caused by the disease, and S represents a symptom caused by the pathological state. We assume that S and D are conditionally independent given P. The potential $\rho$ is the prior probability of D, the potential $\nu$ is the conditional probability of P given D, and the potential $\mu$ is the conditional probability of S given P. A medical intern first observes the symptom S and then either treats the patient for the disease and pathological state or not. The utility valuation $\pi$ bears on the intern's action T, the pathological state P, and the disease D.

Figure 3 shows the results of the fusion algorithm for this for $h_i$, and $\pi_j$ is a utility valuation for $h_j$, then $\pi_i \otimes \pi_j$ is a utility valuation for $h_i \cup h_j$ defined by $(\pi_i \otimes \pi_j)(x) = \pi_i(x^{\downarrow h_i})\pi_j(x^{\downarrow h_j})$ for all $x \in \mathcal{W}_{h_i \cup h_j}$. This method does not apply directly in problems where the utility valuation decomposes additively. In such problems, we first have to combine all utility valuations before we apply the method described in this section. Thus the fusion method described in this section is unable to take computational advantage of an additive decomposition of the utility valuation. In Shenoy [1990b], we describe a modification of the fusion method that is able to take advantage of an additive decomposition of the utility function. The modification involves some divisions.

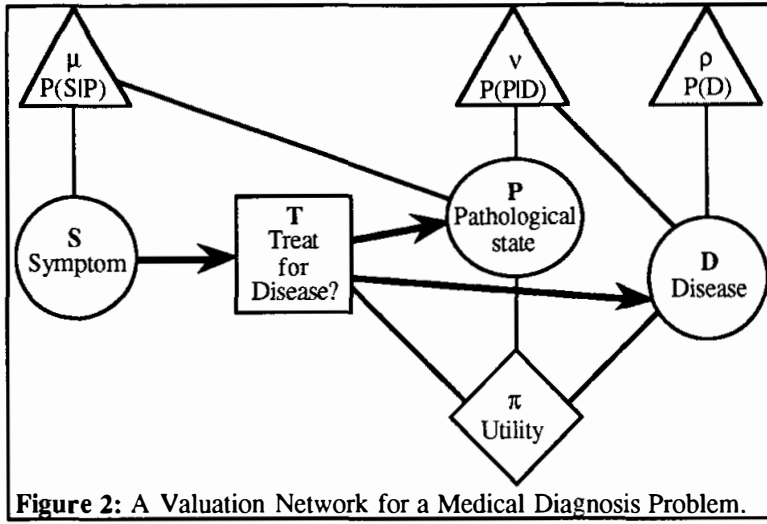

Figure 2: A Valuation Network for a Medical Diagnosis Problem.

problem. The deletion sequence used is DPTS. The first network in Figure 3 is the same as the one in Figure 2. The second network is the result after deletion of D and the resulting fusion. The combination in the fusion operation involves only variables D, P, and T. The third network is the result after deletion of P. The combination operation in the corresponding fusion operation involves only three variables, P, T, and S. The fourth network is the result after deletion of T. There is no combination involved here, only marginalization on the frame of $\{S, T\}$. The fifth network is the result after deletion of S. Again, there is no combination involved here, only marginalization on the frame of $\{S\}$. The maximum expected utility value is given by $((\pi \otimes \nu \otimes \rho)^{\downarrow\{T,P\}} \otimes \mu)^{\downarrow \varnothing}(\blacklozenge)$. An optimal strategy is given by the solution for T with respect to $((\pi \otimes \nu \otimes \rho)^{\downarrow\{T,P\}} \otimes \mu)^{\downarrow\{S,T\}}$, computed during fusion with respect to T. Note that in this problem, the fusion algorithm avoids computation on the frame of all four variables.

In solving the medical diagnosis problem using our method, we do only 9 additions, 20 multiplications and 2 comparisons, for a total of 31 operations. On the other hand, for this problem, decision tree methodology requires 23 additions, 42 multiplications, 12 divisions, and 2 comparisons for a total of 79 operations. Thus, for this problem, our method results in a savings of 61 percent over the decision tree methodology. If we use the influence diagram methodology for this problem, we do 13 additions, 26 multiplications, 8 divisions, and 2 comparisons, for a total of 49 operations. Thus, for this problem, our method results in a savings of 37 percent over the influence diagram methodology.

The fusion method described in this section applies when there is one utility valuation in the VBS. This method applies unchanged in problems where the utility valuation factors multiplicatively into several utility valuations. In this case, we define combination of utility valuations as pointwise multiplication, i.e., if $\pi_i$ is a utility valuation

**Deletion Sequences.** Since > is only a partial order, in general, we may have many deletion sequences (sequences that satisfy the condition stated in Theorem 1). If so, which deletion sequence should one use? First, we note that all deletion sequences lead to the same final result. This is implied in the statement of the theorem. Second, different deletion sequences may involve different computational efforts. For example, consider the VBS shown in Figure 3. In this example, deletion sequence DPTS involves less computational effort than PDTS as the former involves combinations on the frame of three variables only whereas the latter involves combination on the frame of all four variables. Finding an optimal deletion sequence is a secondary optimization problem that has shown to be NP-complete [Arnborg et al., 1987]. But, there are several heuristics for finding good deletion sequences [Kong, 1986; Mellouli, 1987; Zhang, 1988].

One such heuristic is called one-step-look-ahead [Kong, 1986]. This heuristic tells us which variable to delete next from amongst those that qualify. As per this heuristic, the variable that should be deleted next is one that leads to combination over the smallest frame. For example, for the VBS in Figure 5, two variables qualify for first deletion, P and D. This heuristic picks D over P since deletion of P involves combination over the frame of $\{S, D, P, T\}$ whereas deletion of D involves combination over the frame of $\{T, P, D\}$. Thus, this heuristic would choose deletion sequence DPTS.

## 6 CONCLUSIONS

The main objective of this paper is to propose a new method for solving Bayesian decision problems. The VBS representation and solution described here is a hybrid of valuations-based systems for probability propagation [Shenoy, 1991c] and valuation-based systems for optimization [Shenoy, 1991b].

There are several advantages of the VBS representation and solution methodology. First, like influence diagrams, a valuation network representation is compact when compared to decision trees. A valuation network graphically depicts the qualitative structure of the decision problem and de-emphasizes the quantitative details of the problem. However, both VBSs and influence diagrams are appropriate only for symmetric decision problems. For non-symmetric decision problems, decision tree representation is more flexible.

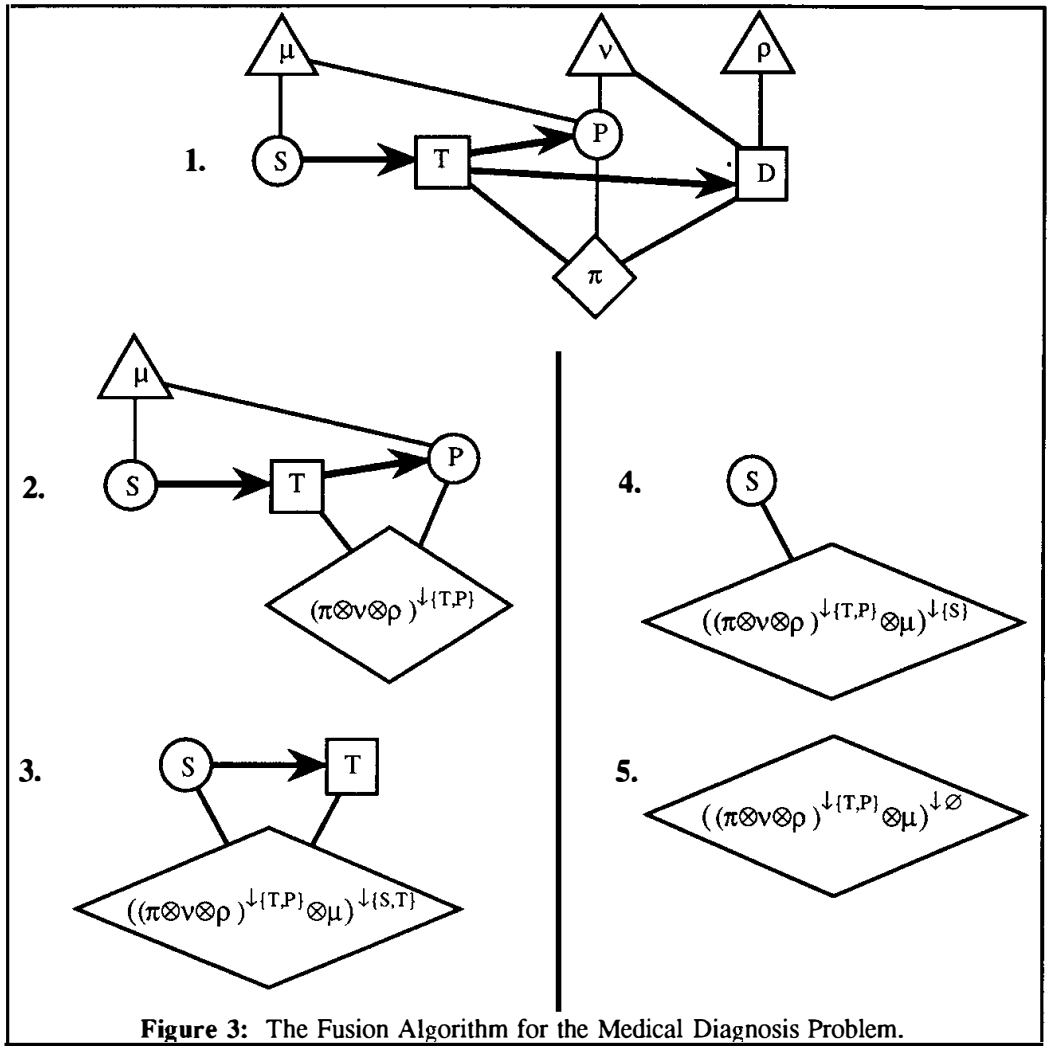

Figure 3: The Fusion Algorithm for the Medical Diagnosis Problem.

Second, like influence diagrams, the VBS representation separates the formulation of the problem from its solution.

Third, in symmetric decision problems, the solution procedure of VBSs is more efficient than that of decision trees since it involves minimal divisions. This assumes that the computational procedure of decision trees includes the preprocessing of probabilities. The solution procedure of decision trees includes unnecessary divisions and multiplications. The unnecessary divisions take place during preprocessing of probabilities. The unnecessary multiplications make up for the unnecessary divisions and take place during the averaging-out process. In non-symmetric decision problems, however, decision trees may be more efficient than VBSs.

Fourth, the VBS representation is more powerful than influence diagram representation. Whereas influence diagram representation is only capable of directly representing conditional probabilities, VBS representation is capable of directly representing arbitrary probabilities. (By directly, we mean without any preprocessing.)

Fifth, the solution method of VBSs involves minimal divisions. In comparison, the influence diagram solution method involves unnecessary divisions (in every arc reversal operation) and additional multiplications to compensate for the unnecessary divisions. These unnecessary divisions and multiplications are the same as those in the decision tree solution process. In influence diagrams, these unnecessary operations are performed for semantical considerations. The influence diagram solution process has the property that the diagram resulting from the deletion of a chance node is again an influence diagram. This means that the resulting probabilities in the reduced influence diagram are conditional probabilities. It is this demand for conditional probabilities at each stage that results in the unnecessary divisions and multiplications.

Sixth, the semantics of VBSs are different from the semantics of influence diagrams. Whereas influence diagrams are based on the semantics of conditional independence, VBSs are based on the semantics of factorization.

Seventh, if a decision problem has no random variables, it reduces to an optimization problem. And the solution technique of VBSs reduces to dynamic programming [Shenoy, 1991b].

Eighth, in cases where a decision problem has no decision variables, we may be interested in finding marginals of the joint distribution for each random variable. In such problems, the solution technique described in this paper reduces to the technique for finding marginals [Shenoy, 1991c]. This technique also can revise marginals in light of new observations. We represent each new observation by a potential and then use the fusion algorithm to compute the desired marginals.

## Acknowledgements

This research was partially supported by NSF grant IRI-8902444. I have benefitted from discussions with and comments from Dan Geiger, Steffen Lauritzen, Anthony Neugebauer, Pierre Ndilikilikesha, Geoff Schemmel, Glenn Shafer, Philippe Smets, Leen-Kiat Soh, Po-Lung Yu, and Lianwen Zhang.This research was partially supported by NSF grant IRI-8902444. I have benefitted from discussions with and comments from Dan Geiger, Steffen Lauritzen, Anthony Neugebauer, Pierre Ndilikilikesha, Geoff Schemmel, Glenn Shafer, Philippe Smets, Leen-Kiat Soh, Po-Lung Yu, and Lianwen Zhang.